\documentclass[9pt, conference]{IEEEtran}
\IEEEoverridecommandlockouts
\usepackage{cite}
\usepackage{amsmath,amssymb,amsfonts}
\usepackage{algorithmic}
\usepackage{graphicx}
\usepackage{textcomp}
\usepackage{multicol}
\usepackage{multirow}
\usepackage{xcolor}
\usepackage{pifont}
\newcommand{\cmark}{\ding{51}}%
\newcommand{\xmark}{\ding{55}}%
\def\BibTeX{{\rm B\kern-.05em{\sc i\kern-.025em b}\kern-.08em
    T\kern-.1667em\lower.7ex\hbox{E}\kern-.125emX}}
\begin{document}

\title{Exploring SSL Discrete Tokens for Multilingual ASR\\

}
\author{
\IEEEauthorblockN{Mingyu Cui$^{1 \dagger }$\thanks{$^{\dagger}$Work done during an internship at Noah's Ark Lab}, Daxin Tan$^{3}$, Yifan Yang$^{2}$, Dingdong Wang$^{1}$, Huimeng Wang$^{1}$, \\ Xiao Chen$^{3}$, Xie Chen$^{2}$, Xunying Liu$^{1}$}
\IEEEauthorblockA{\textit{$^{1}$System Engineering and Engineering Management, The Chinese University of Hong Kong, China}\\
\textit{$^{2}$MoE Key Lab of Artificial Intelligence, X-LANCE Lab, Shanghai Jiao Tong University}\\
\textit{$^{3}$Noah's Ark Lab, Hong Kong SAR, China}}
}

\maketitle

\label{0-abstract}
With the advancement of Self-supervised Learning (SSL) in speech-related tasks, there has been growing interest in utilizing discrete tokens generated by SSL for automatic speech recognition (ASR), as they offer faster processing techniques. However, previous studies primarily focused on multilingual ASR with Fbank features or English ASR with discrete tokens, leaving a gap in adapting discrete tokens for multilingual ASR scenarios. This study presents a comprehensive comparison of discrete tokens generated by various leading SSL models across multiple language domains. We aim to explore the performance and efficiency of speech discrete tokens across multiple language domains for both monolingual and multilingual ASR scenarios. Experimental results demonstrate that discrete tokens achieve comparable results against systems trained on Fbank features in ASR tasks across seven language domains with an average word error rate (WER) reduction of 0.31\% and 1.76\% absolute (2.80\% and 15.70\% relative) on dev and test sets respectively, with particularly WER reduction of 6.82\% absolute (41.48\% relative) on the Polish test set. 

\begin{figure*}[htbp]
\centerline{\includegraphics[width=7.0in]{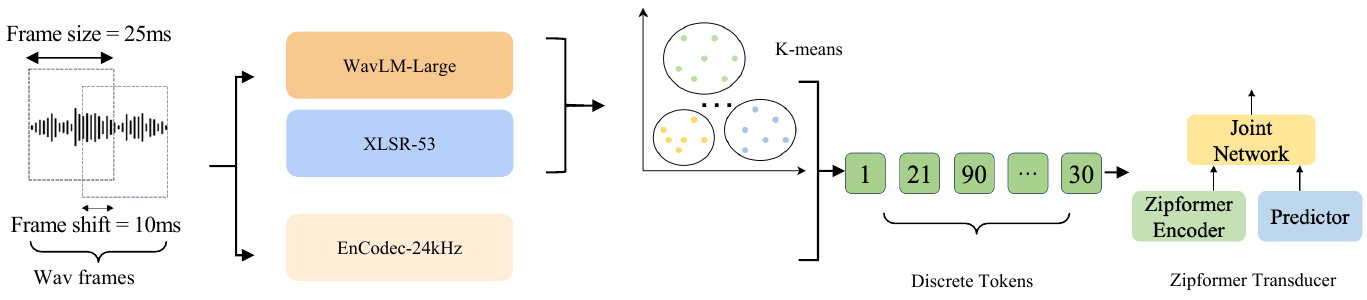}}
\caption{Illustration of the pipeline for discrete speech tokenization, showcasing how a waveform is transformed into discrete tokens. This is achieved either through k-means clustering of XLSR-53/WavLM-Large representations or directly via EnCodec-24kHz quantization}
\label{pipeline}
\end{figure*}
\section{Introduction}
Self-supervised learning (SSL) based speech foundation models \cite{mohamed2022self,chen2022wavlm, baevski2020wav2vec, baevski2019vq}, constructed using large quantities of unlabelled data, have emerged as a powerful paradigm for many downstream speech processing tasks, including automatic speech recognition (ASR). Moreover, discrete tokens generated by SSL models have demonstrated remarkable performance compared to traditional FBank-based models in ASR tasks \cite{graves2006connectionist, ghodsi2020rnn, gulati2020conformer,tsunoo2019transformer, chen2021developing, liu2013use,cui23_interspeech,deng2022confidence,dai2019transformer}. These tokens can be broadly categorized into two main types: semantic tokens and acoustic tokens. Semantic tokens are typically generated by SSL models employing quantization techniques, including vq-wav2vec\cite{baevski2019vq}, wav2vec 2.0\cite{baevski2020wav2vec}, HuBERT\cite{hsu2021hubert}, and WavLM\cite{chen2022wavlm}. These models commonly utilize k-means clustering or Gumbel-Softmax vectorization for quantization and are trained using discrimination tasks or masked prediction objectives. On the other hand, acoustic tokens generated by audio neural codec models like Soundstream\cite{zeghidour2021soundstream} and EnCodec\cite{defossez2022high} focus on accurately reconstructing speech and encoding detailed acoustic information.

SSL discrete speech representations extracted from these models \cite{chang2023exploring, baevski2019effectiveness, chang2023exploration, yang2023towards, chang2024interspeech} are highly adaptable across different task domains including ASR and text-to-speech (TTS) tasks. However, the application of discrete tokens produced by SSL models has predominantly focused on English-related domains, with models such as HuBERT and WavLM demonstrating exceptional performance in English corpora \cite{yang2023towards, chang2023exploration,chang2024interspeech, ueno2021data}. While these studies have laid the groundwork for the utilization of discrete tokens in ASR, their limitation lies in the insufficient exploration of their potential in multilingual scenarios.

In multilingual ASR scenarios, existing research has primarily concentrated on FBank features as input. These approaches, which employ mel-spectral features for direct acoustic modeling, have achieved notable results in multilingual ASR tasks \cite{li2021scaling, pratap2020massively,bai2022joint,li2022recent, shi2024ml}. The successful performance of FBank features can be attributed to their ability to capture the special characteristics of speech across diverse acoustic and linguistic environments. However, there has been limited exploration of the effectiveness and potential of discrete tokens in multilingual ASR tasks across a broader range of languages. 

The under-exploration of discrete tokens in multilingual ASR presents new challenges and benefits. This situation raises questions about the generalizability of discrete token approaches beyond the English language domain. Simultaneously, the unique properties of discrete tokens, which allow them to capture acoustic and semantic information in a compact form, may offer advantages in modeling the diverse phonetic and prosodic features encountered in multilingual scenarios.

To address these challenges and explore the potential benefits, this study aims to investigate the universality and performance of speech discrete tokens across multiple language domains for both monolingual and multilingual ASR scenarios. As illustrated in Fig. \ref{pipeline}, we convert raw audio into speech discrete tokens using three leading speech foundation models: WavLM-Large, XLSR-53, and EnCodec. We then utilize these discrete tokens to train end-to-end (E2E) ASR models for monolingual and multilingual ASR tasks. To fully explore the properties of discrete tokens, we extensively evaluate the efficacy of replacing Fbank-based features with SSL discrete tokens across seven language domains (German (DE), French (FR), Spanish (ES), Dutch (NL), Italian (IT), Portuguese (PT), and Polish (PL)) using the 6000-hour Multilingual Librispeech corpus. Our experiments cover both monolingual and multilingual ASR tasks. Notably, the best-performing Zipformer-Transducer system using discrete tokens outperforms the Fbank-based baselines with an average word error rate (WER) reduction of 0.31\% and 1.76\% absolute (2.80\% and 15.70\% relative) on dev and test sets across the seven languages and shows a remarkable performance with a 6.82\% absolute (41.68\% relative) WER reduction on the Polish test set.

The main contributions of this paper are summarized as follows: 

1) To the best of our knowledge, this work pioneers the use of SSL discrete speech tokens for both monolingual and multilingual non-English language scenarios, extending beyond the English-centric focus of prior research. 

2) We comprehensively demonstrate the effectiveness and efficiency of replacing Fbank features with SSL discrete tokens across seven diverse languages, totaling 6000 hours of speech data from the Multilingual Librispeech corpus.

3) By exploring the properties of discrete tokens in various language contexts, we demonstrate their versatility and robustness, providing valuable insights for developing more inclusive and efficient multilingual ASR systems.

\section{Zipformer-Transducer ASR Architecture}

This paper utilizes the neural Transducer \cite{graves2012sequence} model to perform speech recognition, which is composed of three modules: audio \textbf{Encoder}, text \textbf{Predictor} and \textbf{Joint Network} respectively, as depicted in Fig. 1.  Here we denote  $\mathbf{{x}}_{1:T_i}^{i} = [x^i_1,x^i_2,\cdots, x^i_{T_i}]$ of length $T_i$ as discrete token sequence and $\mathbf{y}_{1:U_i}^{i}=[\mathbf{y}_{1}^{i}, \mathbf{y}_{2}^{i},\cdots,\mathbf{y}_{U_i}^{i}]$ as the corresponding label of length $U_i$. Note that each element in the discrete token sequence $\mathbf{{x}}_{1:T_i}^{i}$ is an integer (codebook index) rather than a vector. Then it is fed into the encoder to produce the acoustic representation $\mathbf{{h}}_{1:{T_i}}^{i}$. The history output labels $\mathbf{{y}}_{1:u-1}^{i}$ are fed into the predictor module to generate the text representation $\mathbf{f}_{u-1}^{i}$. The outputs of the encoder and predictor are then combined in the Joint Network via a non-linear function such as ReLU to obtain the hidden state $\mathbf{g}_{t, u-1}^{i}$ at time step $t$ with output history $\mathbf{y}_{1:u-1}^{i}$. These operations are as follows,

\begin{equation}
    \begin{aligned}
        \mathbf{h}_{1:T_i}^{i} &= \mathrm{Encoder}(\mathbf{x}_{1:{T_i}}^{i}) \\ 
        \mathbf{f}_{u-1}^{i} &= \mathrm{Predictor}(\mathbf{y}_{1:u-1}^{i}) \\
        \mathbf{g}_{t,u-1}^{i} &= \mathrm{Relu}(\mathbf{h}_{1:T_i}^{i} + \mathbf{f}_{u-1}^{i}) \\
        P(\mathbf{y}_{t}^{i}| \mathbf{y}_{1:u-1}^{i}, \mathbf{x}_{1:{T_i}}^{i}) &= \mathrm{Softmax}({\mathbf W}_{o} * \mathbf{g}_{t,u-1}^{i})
    \end{aligned}
\end{equation}
where $\mathbf{W}_{o}$ is a linear transformation applied prior to the final Softmax output layer. Among existing neural Transducer systems, Recurrent Neural Network (RNN) or long short term memory (LSTM) \cite{graves2012sequence, hou2022bring} and Transformer \cite{chen2021developing, zhang2020transformer} architectures have been used for the encoder, while the predictor module is commonly based on LSTM. In this paper, Zipformer-Transducer designed using Zipformer based encoder and Stateless \cite{ghodsi2020rnn} Prediction modules are used throughout this paper.

In contrast to Conformer, the Zipformer\cite{yao2023zipformer} encoder works with the sequence at a steady frame rate. Zipformer adopts a structure akin to U-Net with multiple stacks that each downscale the sequence to various lower frame rates. Second, the block structure is redesigned with two Conformer blocks and additional modules, and the attention weights are reused for efficiency. Additionally, it also proposes BiasNorm as a simpler replacement for LayerNorm, which allows for retaining length information in normalization. To achieve better results, it also replaces Swish with its new activation functions SwooshR and SwooshL. Furthermore, it also develops ScaledAdam, a parameter-scale-invariant version of Adam. In particular, the Zipformer block consists of six modules: Feed-Forward Network (FFN), Non-Linear Attention (NLA), Multi-Head Self-Attention (MHSA), Multi-Head Attention Weights (MHAW), Convolution (CONV), and Forward Network (FFN).
\section{Speech Recognition with discrete tokens}
\subsection{ASR with Discrete Tokens Input}
We explore the application of discrete tokens generated by three SSL models to train E2E ASR systems. Our primary approach employs the XLSR-53 model for speech discretization because it is trained across 53 languages. For comparison, we also examine the discrete token generated by WavLM-Large and EnCodec 24kHz models. An illustration of the pipeline for discrete speech tokenization is shown in Fig. \ref{pipeline}.

Discretization methodology for XLSR-53 and WavLM-Large involves extracting hidden embeddings from the 21st Transformer encoder layer, followed by k-means clustering. The k-means clustering approach is selected for discretizing continuous speech embeddings into discrete token labels for ASR modeling. Moreover, we fine-tuned XLSR-53 for the seven language domains before extracting discrete tokens. In contrast,  EnCodec employs 8 codebooks with 1024 entries to directly output quantized labels. 

For ASR model training, we employ the Zipformer-Transducer architecture, utilizing the discrete speech tokens and their corresponding transcriptions as input. The training process incorporates Recurrent Neural Network Transducer (RNN-T) loss. A linear embedding layer projects the discrete tokens to 80-dimensional vectors. In cases of multiple feature groups, these are embedded, concatenated, and projected to maintain consistent input dimensions. Next, these features are interpolated to a consistent rate of 100 Hz before being input into the ASR model.

\subsection{Discrete Data Augmentation Techniques}
To enhance model robustness and generalization, we implement four data augmentation techniques: 

\textbf{Time Warping}: adjusts the sequence using the \textit{interpolate} function to create variations in timing. A random center within a certain range is selected to warp different segments of the sequence.

\textbf{Time Masking:} hides consecutive parts of the sequence to encourage the model to learn from other parts. The number of masked regions and the width of each mask are randomly determined, introducing randomness and promoting robustness in learning.

\textbf{Embedding Masking:} obscures parts of the vertical embedding dimension, helping the model focus on different aspects of the data. By applying this masking technique twice independently, the model gains additional exposure to varied information.

\textbf{Gaussian Noise:} sampled from a standard normal distribution, is added to the data with a certain probability.

These augmentation strategies play a crucial role in improving the model's ability to handle varied speech inputs which can be found in more detail in \cite{yang2023towards} 
\label{3-experiment}

\begin{table*}[htbp]
\centering
\caption{Monolingual ASR Performance Comparison of Using Fbank and Discrete Token Features as Inputs on Multilingual LibriSpeech}
\label{table_mono}
\scalebox{0.83}{
\begin{tabular}{cccccccccccc}
\hline
\multirow{3}{*}{\textbf{ID}} & \multirow{3}{*}{\textbf{Model}} & \multirow{3}{*}{\textbf{Feature}} & \multirow{3}{*}{\textbf{Units}} & \multicolumn{8}{c}{\multirow{2}{*}{\textbf{WER (\%) dev / test}}}  \\
&&&&&&&\\
\cline{5-12}
 &  &  &  & \textbf{DE} & \textbf{NL}& \textbf{FR} & \textbf{ES}  & \textbf{IT} & \textbf{PT}& \textbf{PL} & \textbf{Avg.} \\
\hline
1 & Whisper-L & \multirow{3}{*}{Fbank} & \multirow{3}{*}{-} & 8.39 / 8.58 & 16.73 / 11.83&10.65 / 8.95 &  6.32 / 5.72& 12.85 / 12.36 &  \textbf{13.26} / \textbf{12.29}&\textbf{10.11} / \textbf{7.38}  & 11.18 / 9.59\\
2 & XLSR-53-CTC & &  &  7.59 / 8.90& 16.10 / 12.46& 12.38 / 10.55&  6.64 / 6.73&13.98 / 12.25 &  19.60 / 18.42&  10.20 / 8.82 & 12.36 / 11.16\\

3 & Zipformer-Transducer &  &  & 4.00 / 5.06& 16.21 / 15.80& \textbf{7.00} / \textbf{5.80}& 5.16 / 5.57& 12.87 / 11.30&19.50 / 18.58 & 11.24 / 16.36 & 10.85 / 11.21\\
\hline


4 &\multirow{3}{*}{Zipformer-Transducer} & Encodec & 1024$^{8}$ & 6.07 / 7.56& 16.56 / 16.20& 7.50 / 7.32&6.27 / 6.89 &14.02 / 12.58 & 19.90 / 18.80& 11.50 / 16.57 & 11.69 / 12.27\\
5 & &Wavlm & 2000 & 4.88 / 5.96 &18.67 / 13.28& 7.92 / 6.58 &5.40 / 6.38&14.46 / 11.66 & 19.47 / 19.27& 11.78 / 12.50 & 11.79 / 10.80\\
6 &  & XLSR-53 & 2000 &\textbf{4.00} / \textbf{5.03}  & \textbf{15.06} / \textbf{11.71}&7.20 / 6.06 &\textbf{4.34} / \textbf{5.54}&  \textbf{12.85} / \textbf{11.10}& 19.15 / 17.21& 11.21 / 9.54 & \textbf{10.54} / \textbf{9.45}\\

\hline
\end{tabular}}

\end{table*}
\label{duration}
\begin{table}[htbp]
    \centering
    \caption{Statistics of Train/Dev/Test partition of each language.}
    \begin{tabular}{c|c|c|c}
    \hline
    \multirow{2}{*}{\textbf{Language}} & \multicolumn{3}{c}{\textbf{Durations (hrs)}} \\
    \cline{2-4}
    &train & dev& test \\ 
    \hline
         German & 1966.51 & 14.28 & 14.29  \\
         Dutch & 1544.24 & 12.76 & 12.76 \\
         French & 1076.58 & 10.07 & 10.07 \\
         Spanish & 917.68 & 9.99 & 10.0 \\
         Italian & 247.38&5.18  &5.27 \\
         Portuguese & 160.96&3.64 &3.74 \\
         Polish & 103.65&2.08 &2.14 \\
         \hline
    \end{tabular}
    
    \label{tab:hours}
\end{table}
\section{Experiments}
\subsection{General Setup}
The efficacy of the proposed pipeline in ASR is assessed on Multilingual Librispeech \cite{pratap2020mls} corpora, including Word Error Rate (WER) for German (DE),  Dutch (NL), French (FR), Spanish (ES), Italian (IT), Portuguese (PT), and Polish (PL) on dev and test sets in Table I. Separate k-means models are trained for each language domain on a 100-hour, randomly selected subset of speech.

In Fbank-based experiments, SpecAugment \cite{park2019specaugment} is applied during training for robustness. The input is 80-channel Fbank features extracted over windows size 25ms with a 10ms frame shift. 500-Byte Pair Encoding (BPE) is applied for the monolingual classification units while 3500-BPE is utilized for multilingual speech training. In discrete tokens experiments, data augmentation is utilized for training as described in Sec. III.B

The Zipformer Transducer \cite{yao2023zipformer} architecture is adopted for ASR implemented with the k2 and icefall\footnote{https://github.com/k2-fsa/icefall/tree/master} framework. The encoder employs a 6-stack Zipformer with downsampling factors of (1,2,4,8,4,2). The label decoder employs a stateless decoder consisting of an embedding layer followed by a 512-dim Conv1D layer. The model has 65.5M parameters. For German, Dutch, French, and Spanish each totaling duration close to or bigger than 1000h, the models are trained for 40 epochs with a learning rate of $10000 / \text{total dataset duration}$. For Italian, Portuguese, and Polish each duration less than 1000h, the models are trained for 150 epochs with a learning rate $10000 / \text{total dataset duration}$. 

The Whisper-Large\footnote{https://huggingface.co/openai/whisper-large-v3}\cite{radford2023robust} model and finetuned XLSR-53-CTC\footnote{https://huggingface.co/facebook/wav2vec2-large-xlsr-53} model\cite{conneau2020unsupervised} are utilized for ASR training with FBank input. Moreover, the discrete tokens are extracted from the 21-st layer of XLSR-53 with 2000 units, the 21-st layer of WavLM large\footnote{https://huggingface.co/microsoft/wavlm-large}\cite{chen2022wavlm} with 2000 units, and EnCodec\footnote{https://huggingface.co/facebook/encodec\_24khz}\cite{defossez2022high} 24kHz with $1024^8$ units.

\subsection{WER Results on Multilingual LibriSpeech}
TABLE I presents a WER performance comparison of models trained on individual languages from the Multilingual LibriSpeech corpus, totaling 6000 hours, using various discrete token inputs. Sys. 1-3 comprise Whisper-Large, fine-tuned XLSR-53-CTC on each monolingual language, and Zipformer-Transducer models with Fbank features as input. Sys. 4-6 are Zipformer-Transducer models utilizing different discrete tokens generated by EnCodec-24kHz, WavLM-Large, and fine-tuned XLSR-53-CTC. Several trends can be found:

1) Among the experiments with Fbank features as input (Sys. 1-3, TABLE I), Zipformer-Transducer achieves almost the best or near-best performance across all monolingual languages except Polish and Portuguese. Therefore, we utilize Zipformer-Transducer architecture in discrete token experiments, while three systems (Sys.1-3, TABLE I) with Fbank feature serve as baseline systems for comparison.

2) The Zipformer-Transducer model utilizing XLSR-53 discrete tokens as input (Sys. 6) outperforms its counterpart using Fbank features (Sys. 3) across all monolingual languages, with the sole exception of French. This superior performance is evident in the substantial WER reduction observed: the best performance showed a remarkable performance with a 6.82\% absolute (41.68\% relative) WER reduction on the Polish test set and an average WER reduction of 0.31\% and 1.76\% absolute (2.80\% and 15.70\%) on the dev and test sets across seven domains. 

3) The Zipformer-Transducer model utilizing XLSR-53 discrete tokens as input (Sys 6) outperforms all models using Fbank features (Sys. 1-3) across most languages, with the exceptions of Portuguese and Polish. Especially, the best-performing systems demonstrate significant WER reduction 4.39\% and 3.55\% absolute (52.62\% and 41.37\% relative) on German dev and test sets (Sys.6 vs. Sys.1).

4) The Zipformer-Transducer utilizing EnCodec discrete tokens (Sys. 4) exhibits suboptimal performance, possibly attributed to the EnCodec's encoding of all speech aspects without disentanglement. In contrast, XLSR-53 and WavLM prioritize speech semantics and effectively support the ASR task.

5) Additionally, the Zipformer-Transducer with WavLM-Large discrete token (Sys. 5) demonstrates inferior results, which can be attributed to the linguistic limitations of WavLM's training data. While WavLM was primarily trained on English data, XLSR-53 benefited from exposure to 53 languages during its training process. Consequently, for non-English speech recognition tasks, WavLM struggles to generate precise and relevant discrete tokens, hindering its effectiveness in multilingual ASR applications. Thus, we can infer that the WavLM pretrained on the English domain may struggle to transfer its self-supervised learning capabilities effectively to low-resource languages that it has not encountered before.
\label{table1}
\begin{table*}[htbp]
\caption{Ablation Study: Comparison of Monolingual vs. Multilingual Training with Discrete Tokens on Multilingual LibriSpeech.}
    \centering
    \scalebox{0.80}{
    \begin{tabular}{cccccccccccc}
    \hline
\multirow{3}{*}{\textbf{ID}}  & \multirow{3}{*}{\textbf{Mix data}} & \multirow{3}{*}{\textbf{Shared Kmeans}} & \multirow{3}{*}{\textbf{Units}} & \multicolumn{7}{c}{\multirow{2}{*}{\textbf{WER (\%) dev / test}}} \\
   &  &  &  & \multicolumn{7}{c}{} \\
\cline{6-12}
 &    &  &  & \textbf{DE} & \textbf{NL} & \textbf{FR} & \textbf{ES} & \textbf{IT} & \textbf{PT} & \textbf{PL} & \textbf{Avg.} \\

 \hline

 \hline
  1 & \xmark & \xmark &2000&\textbf{4.00} / \textbf{5.03}  & 15.06 / \textbf{11.71}&\textbf{7.20} / \textbf{6.06} &\textbf{4.34} / 5.54&  \textbf{12.85} / \textbf{11.10}& 19.15 / 17.21& \textbf{11.21} / \textbf{9.54} & \textbf{10.54} / \textbf{9.45}\\

   2 & \xmark & \cmark &2000& 4.35 / 5.41& 14.96 / 11.82& 8.41 / 6.88 & 5.11 / 6.35& 13.30 / 11.70& 20.28 / 19.27& 17.03 / 16.59 &11.92 / 11.14\\
   3 & \xmark & \cmark & 4000 &4.11 / 5.23 & 14.93 / 11.76 & 8.00 / 6.53& 4.78 / 5.80 &13.23 / 11.59 & 18.90 / 18.32& 13.27 / 13.16 & 11.03 / 10.34\\
    4 & \cmark & \cmark &4000&4.00 / 5.04&\textbf{14.79} / 12.03 &7.94 / 6.35 & 4.40 / \textbf{5.42} &12.74 / 11.01&\textbf{18.69} / \textbf{17.94}& 12.78 / 11.90 & 10.76 / 9.95\\
    
\hline
         
    \end{tabular}}
\end{table*}

\subsection{Efficiency Analysis  of Discrete Tokens}

\begin{figure}[htbp]
\centerline{\includegraphics[width=2.7in]{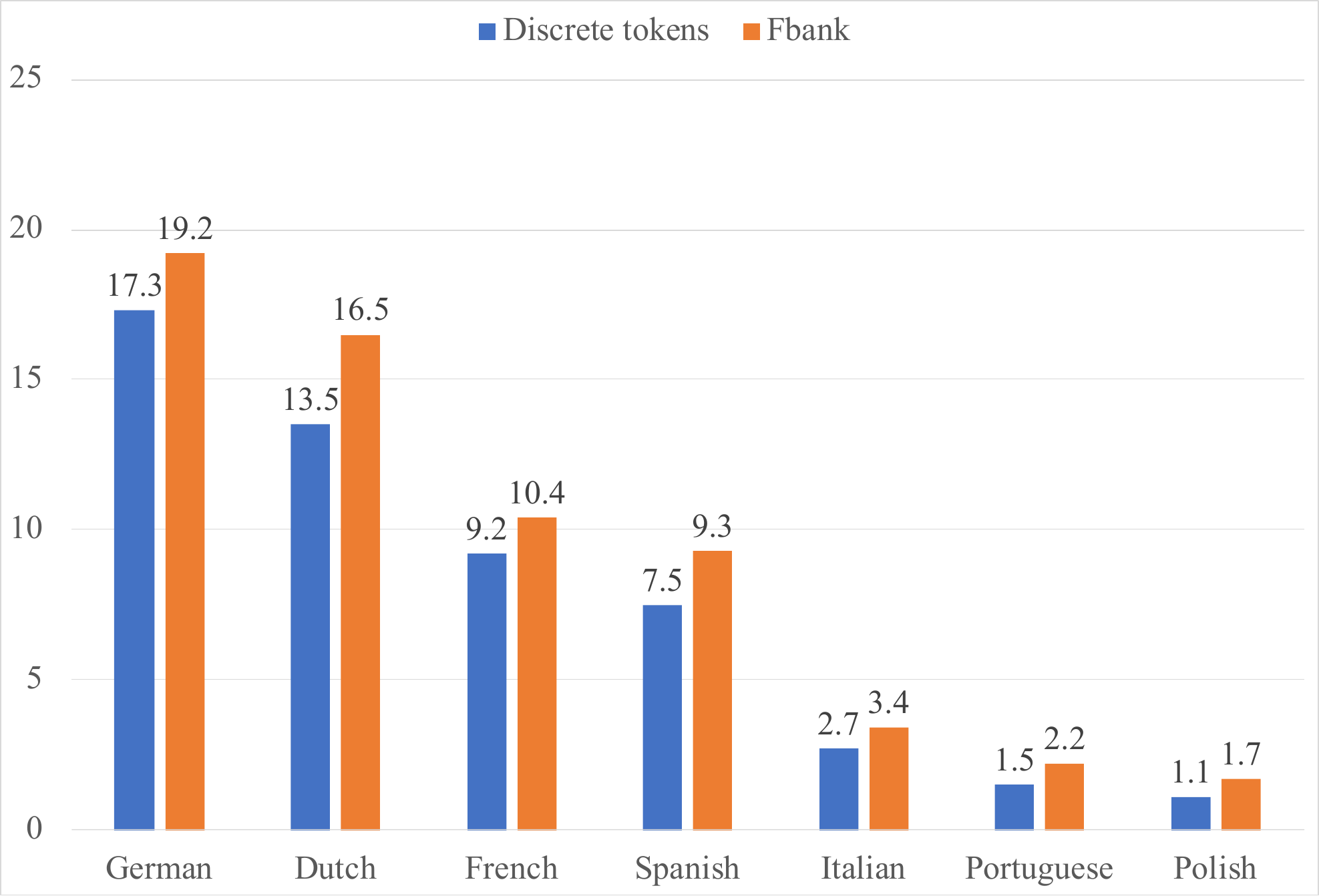}}
\caption{Illustration of training time (minutes) per epoch using Discrete tokens / Fbank features.}
\label{fig}
\end{figure}

Figure \ref{fig} presents a comparative analysis of training durations (minutes) for one epoch across diverse language domains, contrasting the use of FBank features and discrete tokens. To ensure a fair comparison of training efficiency between Fbank and discrete token approaches, both experiments were conducted with an equal number of epochs. The results demonstrate that training with discrete tokens yields significantly reduced training times across all languages examined. This efficiency gain is particularly pronounced in the Polish language domain, where the training time using discrete tokens is remarkably less than 35\% of the time required when utilizing FBank as input features.

This substantial reduction in training time can be attributed to the compact representation provided by discrete tokens, which effectively condense the rich acoustic information present in speech signals. Such compression not only accelerates the training process but also potentially reduces computational resource requirements. These findings suggest that the adoption of discrete tokens in multilingual speech recognition systems could lead to more efficient model development and deployment, especially in scenarios where computational resources are constrained or rapid iteration is imperative.
\vspace{-0.3cm}
\subsection{Ablation Study}
To evaluate the multilingual capabilities of discrete tokens, we examine the data mixing, the effects of shared k-means, and k-means units across seven language domains. Data mixing is the process of combining seven language domains into a unified dataset. The shared k-means was utilized to extract discrete tokens across all seven languages, allowing us to assess the effectiveness of this unified tokenization method in a multilingual context. To better compare the performance of discrete tokens, we use the best setup from the monolingual experiment of XLSR-53 (Sys.6, TABLE I, also showing Sys.1 in TABLE III) to conduct an ablation study. From the TABLE III, several trends can be found:


1) When comparing multilingual ASR with monolingual ASR (Sys.2 and Sys.3 vs. Sys.1), we observe that the shared k-means approach (using 2000 or 4000 clusters shared across 7 languages) underperforms compared to the monolingual approach (where each language has its own 2000 clusters). This performance gap likely arises from the challenge of adequately representing the nuanced characteristics of all seven languages with a shared set of 2000 or 4000 clusters, whereas 2000 clusters appear sufficient for capturing the features of a single language. As a result, distinct phonetic or semantic elements from different languages may be incorrectly assigned to the same discrete token label in the multilingual model, leading to reduced accuracy.

2) The performance comparison between the shared k-means models with 2000 and 4000 clusters (Sys.2 vs. Sys.3) reveals that a higher cluster count yields better results. This suggests that increasing the number of clusters enables the extraction of more refined discrete tokens, thus capturing subtler speech features. This enhanced granularity in token discretization allows for a more precise encoding of acoustic characteristics but doesn't necessarily outperform the monolingual approach (Sys.1).

3) The Zipformer-Transducer with mixing data and employing a shared k-means model achieves comparable performance to its counterpart without these features (Sys.4 vs. Sys.1) while potentially offering better cross-lingual generalization through its multilingual training strategy. We also attempted to mix data and employ a shared k-means model with 2000 units. However, this system failed to converge, indicating that using a shared K-means model with 2000 units is insufficient for effective training.

4) Across all seven languages, the performance of the multilingual systems (Sys.2-4) consistently underperforms the monolingual training approach (Sys.1). This uniform underperformance strongly indicates that the optimal strategy for multilingual model development is to train separate models for each individual language, rather than employing mixed-language training methods.

\label{4-conclusion}
\section{Conclusion}

Exploring the universality of speech discrete tokens across multilingual and monolingual ASR for seven languages, this paper presents a comprehensive study yields significant insights into the universality and efficacy of these tokens. Through an extensive investigation of discrete tokens derived from three leading SSL models: WavLM, XLSR-53, and EnCodec, experimental results indicate that discrete tokens are competitive with FBank features in ASR tasks across seven language domains in both multilingual and monolingual scenarios. Notably, the best-performing Zipformer-Transducer system using discrete tokens outperforms the Fbank-based baselines with an average WER reduction of 0.31\% and 1.76\% absolute (2.80\% and 15.70\% relative) on the dev and test sets across the seven languages and shows a remarkable performance with a 6.82\% absolute (41.68\% relative) WER reduction on the Polish test set. In terms of training efficacy, training with discrete tokens consistently outperforms FBank features across all languages, with the Polish domain demonstrating a notable reduction in training time to less than 35\% of that required when using FBank features. These empirical findings suggest that universal discrete tokens have considerable potential across diverse multilingual speech tasks. 
We hope that this work serves as the foundation for multilingual ASR, paving the way for future research on discrete tokens in diverse language recognition.

\newpage
\bibliographystyle{IEEEtran}
\bibliography{IEEEfull,mybib}

\end{document}